\documentclass[a4paper,conference]{IEEEtran}

\ifCLASSINFOpdf
  \usepackage[nocompress]{cite}
\else
  \usepackage{cite}
\fi
\usepackage[
  colorlinks=true
]{hyperref}

\usepackage{graphicx} 
\def\BibTeX{{\rm B\kern-.05em{\sc i\kern-.025em b}\kern-.08em
    T\kern-.1667em\lower.7ex\hbox{E}\kern-.125emX}}
\usepackage{amsmath} 

\usepackage{makecell}
\usepackage{comment}
\usepackage{lipsum}
\usepackage{graphicx}
\usepackage{cleveref}
\usepackage{longtable}
\usepackage[super]{nth}
\usepackage{cleveref}
\usepackage[table]{xcolor}
\usepackage{tabu}
\usepackage{multirow}
\usepackage{cellspace, graphicx, makecell}

\newcommand\rowincludegraphics[2][]{\raisebox{-0.45\height}{\includegraphics[#1]{#2}}}

\newcommand{\etal}{\textit{et al.}}

\newcommand{\ie}{\textit{i.e.}}
\begin{document}

\title{Transformer-Encoder Detector Module: Using Context to Improve Robustness to Adversarial Attacks on Object Detection}

\author{\IEEEauthorblockN{Faisal Alamri}
\IEEEauthorblockA{Department of Computer Science\\
The University of Exeter\\
Email: fa269@exeter.ac.uk}
\and
\IEEEauthorblockN{Sinan Kalkan}
\IEEEauthorblockA{Department of Computer Engineering\\
Middle East Technical University\\
Email: skalkan@ceng.metu.edu.tr}
\and
\IEEEauthorblockN{Nicolas Pugeault}
\IEEEauthorblockA{School of Computing Science\\
University of Glasgow\\
Email: nicolas.pugeault@glasgow.ac.uk}
}

\maketitle

\makeatletter{\renewcommand*{\@makefnmark}{}
\footnotetext{
This work is accepted for the 25th International Conference on Pattern Recognition (ICPR'2020).
}\makeatother}

\begin{abstract}
Deep neural network approaches have demonstrated high performance in object recognition (CNN) and detection (Faster-RCNN) tasks, but experiments have shown that such architectures are vulnerable to adversarial attacks (FFF, UAP): low amplitude perturbations, barely perceptible by the human eye, can lead to a drastic reduction in labelling performance. 
This article proposes a new context module, called \textit{Transformer-Encoder Detector Module}, that can be applied to an object detector to (i) improve the labelling of object instances; and (ii) improve the detector's robustness to adversarial attacks. 
The proposed model achieves higher mAP, F1 scores and AUC average score of up to 13\% compared to the baseline Faster-RCNN detector, and an mAP score 8 points higher on images subjected to FFF or UAP attacks due to the inclusion of both contextual and visual features extracted from scene and encoded into the model. 
The result demonstrates that a simple ad-hoc context module can improve the reliability of object detectors significantly. 
\end{abstract}

\IEEEpeerreviewmaketitle

\section{Introduction}

Recognising objects in a visual scene is an effortless task for humans, one that has challenged computer vision since its inception. The advent of deep neural network approaches over the last decades has delivered major improvements on this task on all benchmarks, although some difficulties remain \cite{ZhengxiaDBLP:journals/corr/abs-1905-05055}.



One notable weakness of deep neural network approaches to vision, first demonstrated by Szegedy~\etal~\cite{42503}, is that perturbations of very small amplitude, barely visible to the human eye, can be found that lead to major decrease in the neural networks' labelling accuracy---so called \textit{adversarial attacks}. This seminal finding was confirmed by following studies, which showed that such attacks could be designed to be independent on the specific neural network or even neural architecture employed \cite{MainPaper_Ch5}, and that such attack patterns could be data independent \cite{FastFeatureFool}.

This apparent weakness is in contrast to how robust human perception of objects is to a variety of noise and perturbations. One plausible reason for this difference in robustness between human and artificial perception is \textit{context} \cite{bar2004visual}. 
Contextual information plays an important role in visual recognition for both human and computer vision systems \cite{bar2004visual,rabinovich2007objects}. Imagine there are an \textit{office desktop}, a \textit{chair}, a \textit{PC}, a \textit{monitor} and a \textit{keyboard}, then we ask the question ``\textit{what is the missing object in this scene?}". Some may say a \textit{cup}, or a \textit{pen}, but when looking deeply and understanding the context, you are likely to say the missing object is a \textit{mouse}. Such guesses are made even without seeing or gazing at the scene due to leveraging contextual information. This shows how context carries rich information about visual scenes. In terms of object recognition, the context could be defined as cues captured from a scene that presents knowledge about objects locations, size and object-to-object relationships. Due to the importance of contextual information and how it improves detection, it has been widely studied \cite{bozcan2019cosmo, NNMN7239534, Galleguillos:2010:CBO:1801040.1801854, Tree_Model, Mottaghi_6909514}. 
Contextual information can be captured from images explicitly as attempted in \cite{FaisalAlamri1}, or implicitly as proposed in this paper upon the success of the Transformer model proposed by \cite{Attention:journals/corr/VaswaniSPUJGKP17} for Natural Language Processing.

This paper describes a contextual detection module, proposed as an add-on to classical object detection architectures such as Faster-RCNN, named \textit{Transformer-Encoder Detector Module} (TEDM). This model implicitly encodes contextual statistics of objects and uses attention mechanism to improve the labelling of image regions. It improves the detection performance of a state-of-the-art object detector, as evaluated on natural images and perturbed images. This model does rescore, relabel, and correct detector predictions, in contrast to \cite{Tree_Model, 6891238}, which only rescore detected objects' confidences. 

This paper is organised as follows: First, we review the Transformer Model, and how it works. CNN-based detectors (Section \ref{Setection_2}). In Section \ref{Method}, we explain how the proposed model is built and how features are encoded in the pipeline. Finally, the proposed model is examined on MSCOCO2017 dataset in comparison with Faster RCNN detector and the Relabelling Model proposed by \cite{FaisalAlamri1} in Section \ref{Experiments}. Natural and perturbed images are both used to evaluate the proposed model. 

\section{Related Work} \label{Setection_2}

\subsection{Adversarial attacks}
The vulnerability of deep neural network architectures to small amplitude perturbations optimised to damage the networks' performance has first been found by Szegedy~\etal~\cite{42503}. They demonstrated that image perturbations of small amplitude, barely visible to the eye, when optimised to cause the highest misclassification error, can achieve large decreases in performance. Moreover, such perturbations can decrease performance across network architectures and even on different datasets. 

Mohsen \etal~\cite{MainPaper_Ch5} proposed an approach to find Universal Adversarial Perturbations (UAP), single patterns of perturbation that lead to large misclassification across all inputs of the network, and demonstrated its efficiency on various classifiers (e.g. CaffeNet \cite{Jia:2014:CaffeNet}, VGG-16 \cite{VGG_16}, VGG-19 \cite{VGG19}), claiming that the addition of the perturbations drops the tested classifiers’ performance by almost 90\% as tested on ImageNet validation dataset. Therefore, this leads to the result that the single universal perturbation vectors can cause most of the natural images to be misclassified. Therefore, it can be concluded that using UAP perturbations optimised on a network architecture X (e.g., VGG-F) will also lead to misclassification when applied to images classified with another network architecture Y (e.g., CaffeNet). 

Moreover, Reddy \etal~\cite{FastFeatureFool} claim that their proposed model, named Fast Feature Fool (FFF), is transferable to various different models (i.e. they use the same models examined in \cite{MainPaper_Ch5}). The main and important difference between UAP and FFF is that FFF is data-independent. In other words, it can be applied on any model even if the trained and tested models are trained on different datasets (i.e., no prior knowledge about the target CNNs is needed). It is also reported that when FFF is trained on X dataset and evaluated on Y dataset, it can still impact the CNN models, leading to an increase in the fooling rate compared to UAP, which is a data-dependent method. 

\subsection{Visual context for object detection}
Contextual information can help improve detection performance due to the knowledge it provides from scenes \cite{BestDBLP:journals/corr/abs-1809-02165, Galleguillos:2010:CBO:1801040.1801854}. It is defined in \cite{FaisalAlamri2} as \textit{any data obtained from an object’s own statistical property and/or from its vicinity, including intra-class and inter-class details}. Such a definition is claimed due to the information observed while studying the importance of context in digital images. 

It is said that contextual information is a tool used more with multiple objects so that relationships among objects can be deeply understood \cite{Desai2011}. Roozbeh \etal~\cite{Mottaghi_6909514} also state that in digital images, objects with clear appearance (e.g. large objects) are easy to detect, whereas some small objects are harder. Lubor \etal~\cite{10.1007/978-3-642-15561-1_18} also claim that contextual information, therefore, can be a solution here as it provides stronger cues in detecting small objects due to the context where those objects are present. Hence, contextual information is described as \textit {“a natural way to improve detection”} \cite{ContextStudyDBLP:journals/corr/abs-1711-05705,bozcan2019cosmo}. 

Contextual information in the field of object detection can help to understand and explore object vicinity (\ie, scene-level context) as applied in \cite{DBLP:journals/corr/BellZBG15, Zeng2016GatedBC}, and also provides object-object relationships (\ie, object-level context) as in \cite{Rabinovich2007ObjectsIC, DBLP:journals/corr/ChenG17b}. Moreover, Contextual information has been also studied in different areas, such as object localisation \cite{Tree_Model}, image segmentation \cite{Gould2008}, image annotation \cite{5678649}, scene modeling \cite{bozcan2019cosmo} and cognitive robotics \cite{NNMN7239534}.

Context can be classified upon the sources of information extracted from images. Biederman \etal~\cite{BIEDERMAN1982143} state that there are five categories of object-environment dependencies, which are: \textit{“(i) \textbf{interposition} objects interrupt their background, (ii) \textbf{support}: objects often rest on surfaces, (iii) \textbf{probability}: objects tend to be found in some environments but not others, (iv) \textbf{position}: given an object in a scene, it is often found in some positions but not others, and (v) \textbf{familiar size}: objects have a limited set of sizes relative to other objects”}. Galleguillos \etal~\cite{Galleguillos:2010:CBO:1801040.1801854} grouped those relationships into three main categories: (i) \textbf{Semantic} (Probability), (ii) \textbf{Spatial} (interposition, support and position) and (iii) \textbf{Scale} (familiar size). 

\subsection{Attention and the Transformer model}
\begin{figure} 
\centering
\includegraphics[width=0.35\linewidth]{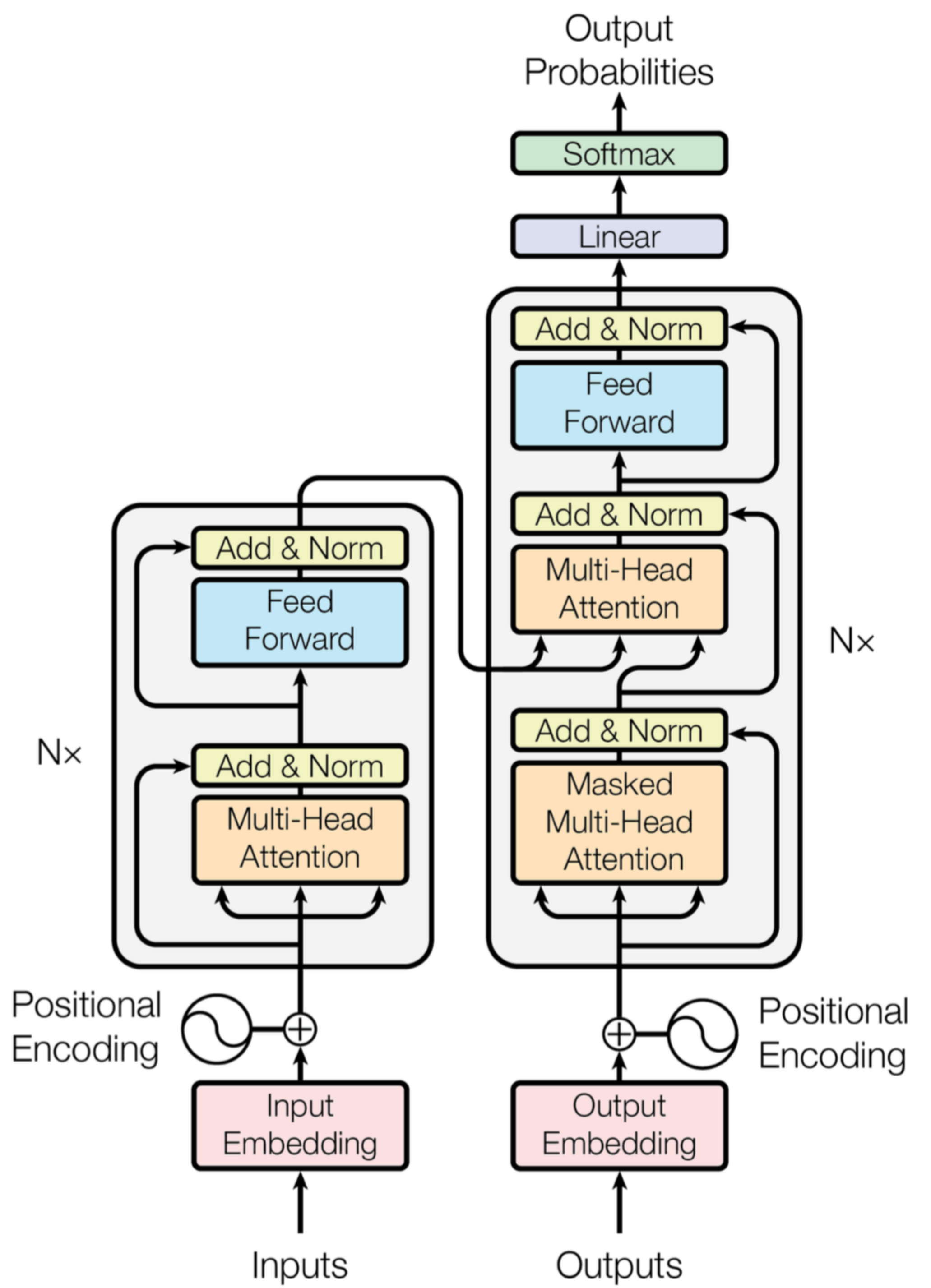}
\caption{The Transformer Architecture, reproduced from \cite{Attention:journals/corr/VaswaniSPUJGKP17}}
\label{fig:TransformerArchitecture}
\end{figure}

Vaswami~\etal~\cite{Attention:journals/corr/VaswaniSPUJGKP17} proposed the Transformer architecture for natural language processing, as a way to encode word context from documents corpora using dot product attention between word vectors.  
The Transformer consists of a set of encoders and decoders, which are composed of a stack of layers. In terms of encoders: each has two major components, which are (1) self-attention layer and (2) feed-forward neural network. On the other hand, the decoder has two similar components and an additional one: (1) self-attention layer, (2) encoder-decoder layer, and (3) feed-forward neural network. The Transformer architecture is shown in Figure \ref{fig:TransformerArchitecture}. We refer the reader to \cite{Attention:journals/corr/VaswaniSPUJGKP17} for further information. 
The Transformer architecture, due to its performance and speed, has been used in a variety of approaches in NLP, such as  \cite{BERT:journals/corr/abs-1810-04805, 8416973, DBLP:journals/corr/abs-1906-08237, Radford2018ImprovingLU}, and has recently been applied to computer vision problems \cite{Carion2020EndtoEndOD}.

\section{Proposed Method} \label{Method}
\begin{figure} 
\centering
\includegraphics[width=0.5\linewidth]{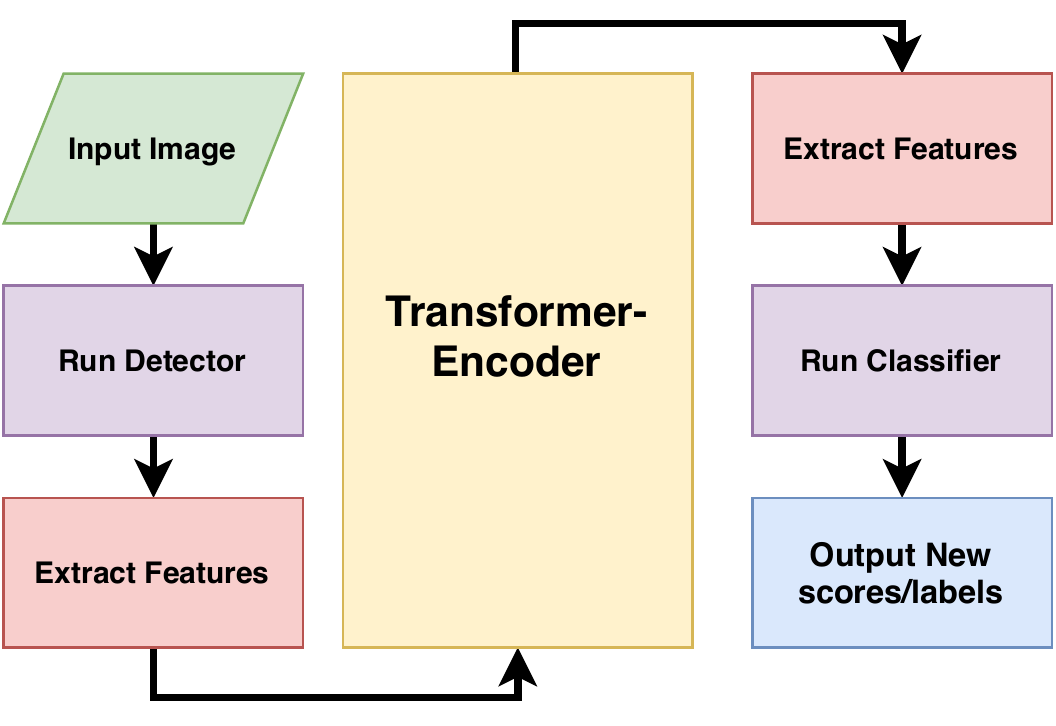}
\caption{Transformer-Encoder Detector Module Architecture}
\label{fig:NewMethodModel}
\end{figure}

A novel model is proposed in this paper, which obtains a better performance compared to the Faster-RCNN detector whether perturbations are applied to the images or not. The proposed model is built upon the success of the Transformer model proposed by \cite{Attention:journals/corr/VaswaniSPUJGKP17}. However, only the Transformer-Encoder is adapted, as shown in Figure \ref{fig:NewMethodModel}, where the proposed model, named \textit{Transformer-Encoder Detector Module} (TEDM) architecture is illustrated. 

First, an image is passed to the baseline detector (i.e., Faster-RCNN) for feature extraction. Faster-RCNN, proposed by Shaoqing \etal~\cite{FasterRCNN:journals/corr/RenHG015}, uses a two stages for detection. In the first stage, it uses a CNN network to extract regions in the image that are likely to contain an object. The second stage, moreover, performs detection (classification \& localisation) for each region proposed by the first stage. Faster-RCNN has been implemented in a variety of articles studying the importance of contextual information (e.g. \cite{Context5DBLP:journals/corr/abs-1711-11575,7952437}), and it is still one of the state-of-the-art detection methods, and due to some of its advantages (e.g. speed, accuracy), it is also used as the baseline detector in this paper.

Features representing image regions for the detected objects are extracted. As the dimension of the features extracted is $4,096$, they are mean-pooled to $2,048$ to suit further processing. The pooled features and the boundary boxes obtained from the detector representing the detected regions are passed into the Transformer-Encoder with a dimension of $2,048$.
Boundary boxes are included to encode the spatial features to extract the contextual information among regions implicitly. Since there is no recurrence in the model, the positions of the inputs are needed, thus positional encoding is applied, which ``injects some information about the relative or absolute position of the tokens in the sequence", using sine and cosine functions of different frequencies. In other words, it is used to obtain the context of each feature inputted \cite{Attention:journals/corr/VaswaniSPUJGKP17}.

The Transformer Model, moreover, is then applied, which is already pre-trained on MS COCO 2014, adapted from \cite{krasserm86:online, Attention:journals/corr/VaswaniSPUJGKP17}. In details, the Transformer-Encoder takes the passed features and processes them with the attention mechanism to output features with a dimension of 512. A feed-forward NN classifier is then applied. The classifier produces the new scores and predicted labels for each region. We use a \textit{trainscg} (scaled conjugate gradient back-propagation) Neural Network approach, as implemented in MATLAB \cite{MATLAB:2017}. 
Scaled conjugate gradient (SCG), a supervised learning algorithm, is a network training function used to update weight and bias value according to the scaled conjugate gradient method \cite{ClassifierMOLLER1993525}. \textit{trainscg} was implemented as explained in \cite{MATLAB:2017}.
The standard network consists of a two-layer feed-forward network, with a sigmoid activation function in the hidden layer, and a softmax function in the output layer. The number of hidden units defined is 200, as different numbers were tested, and 200 achieved the highest performance for this data. 

This method, to the best of our knowledge, is the first to include the Transformer-Encoder to benefit from the use of attention and the positional encoding operation to apply for object detection performance in both natural and perturbed images. The positional encoding, as said, helps to simply encode the semantic and spatial contexts for each region inputted implicitly. \textit{TEDM} is examined, as below, on the MS COCO 2017 \emph{val} dataset in comparison with Faster RCNN (i.e., baseline detector) on two types of images: (1) Natural images, where no perturbations are added, and (2) perturbed images, as presented in Section \ref{Experiments}.

\section{Experiments} \label{Experiments}
This section presents the results obtained from the application of the new proposed method (\textit{TEDM}). Images before any attacks, called natural images, are used to examine the impact of this model in comparison with Faster-RCNN, as in Experiment one (Section \ref{NoAttackSection}). Experiment Two (Section \ref{AttackSection}) illustrates the examination of the \textit{TEDM} vs. Faster RCNN on images under adversarial attacks.

\subsection{Experiment One: Natural Images} \label{NoAttackSection}
In this section, we are presenting how effective \textit{TEDM} is in comparison with Faster RCNN when applied on natural images. 

Note that images used in these experiments are taken from MS COCO 2017 \emph{val} dataset. Only the images with more than one object detected by Faster RCNN are used in order to compare the performance of this proposed model with the \textit{Relabelling Model} proposed by \cite{FaisalAlamri1}, which explicitly extracts contextual features from images. This model improves the detection performance of Faster RCNN detector as it rescores and relabels predictions. using some  Refer to \cite{FaisalAlamri1, FaisalAlamri2} further reading about contextual information and how \textit{Relabelling Model} works.

\begin{figure}
\begin{center} 

\begin{tabular}{|c|c|}  \hline 
{\color{blue}{\cellcolor{green!25}Faster RCNN}}  & {\color{blue}{\cellcolor{green!25}TEDM}} \\ \hline

\includegraphics[width=0.25\linewidth]{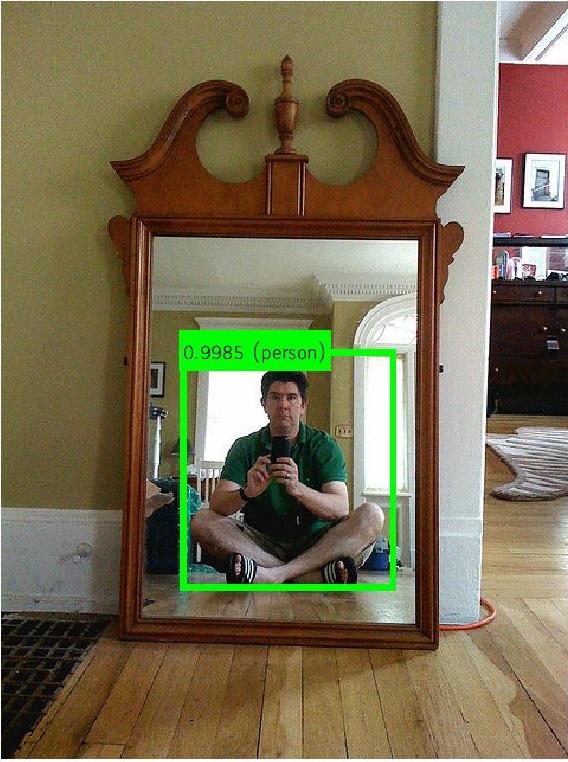}
& 
\includegraphics[width=0.25\linewidth]{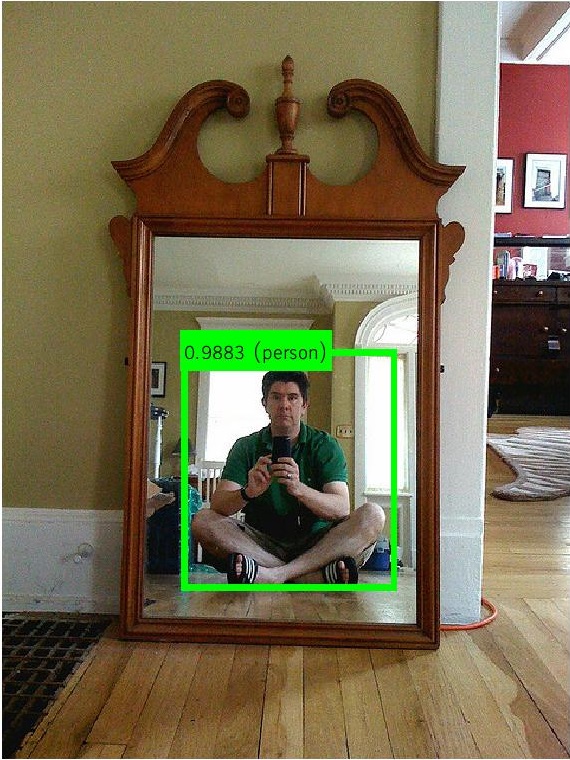}
\\ \hline

\includegraphics[width=0.25\linewidth]{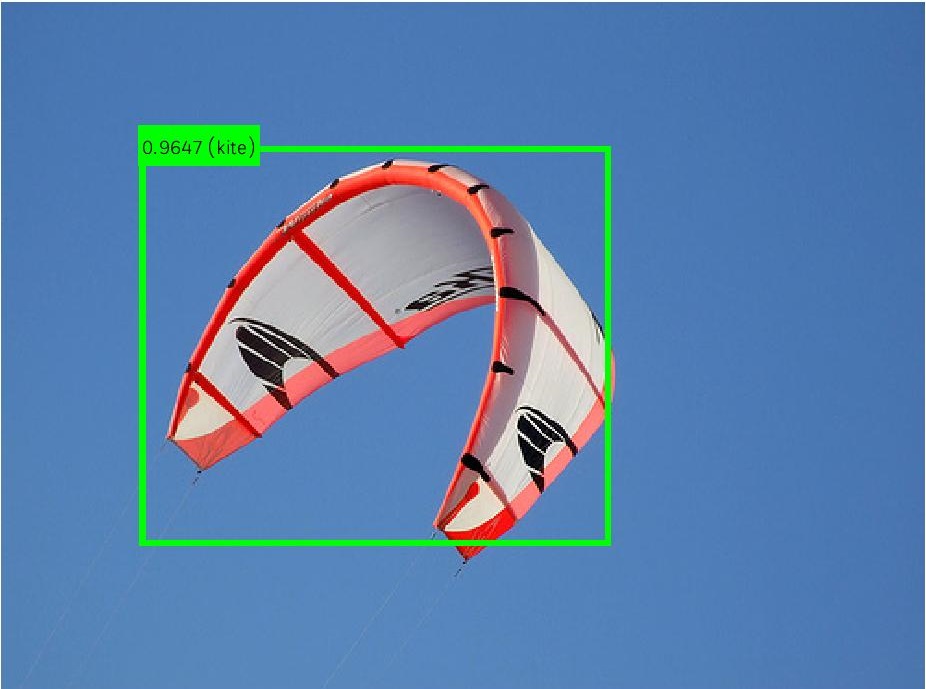}
& 
\includegraphics[width=0.25\linewidth]{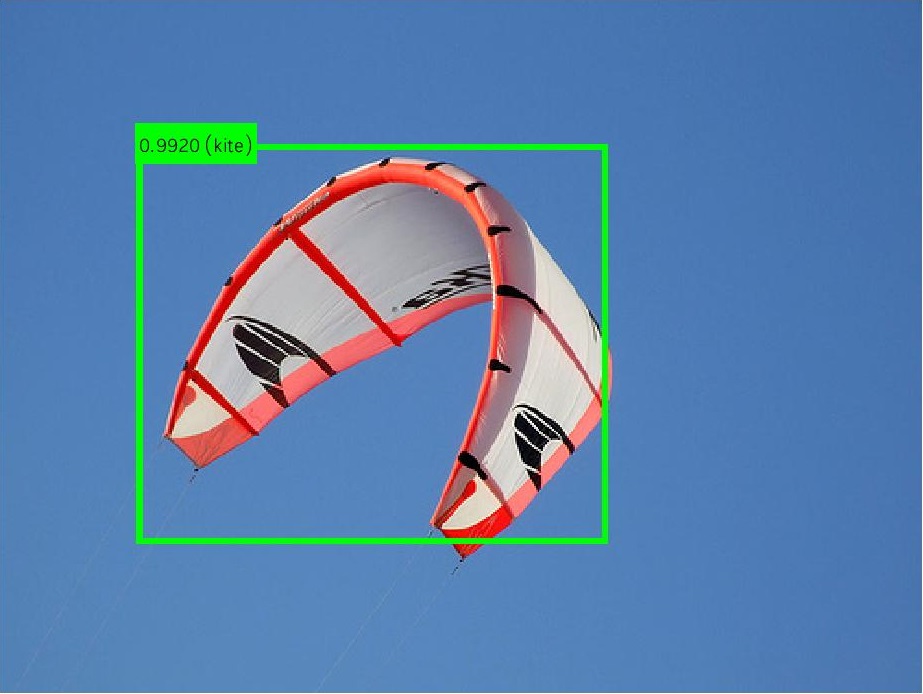}
\\ \hline

\end{tabular}
\end{center}
\caption{Results: Faster RCNN vs. TEDM outputs on images with a single Object: Green boxes represent correct detection.}
\label{fig: singelObject}
\end{figure}

First, two chosen images with only one object presented are used to examine the performance of \textit{TEDM} when a single object is presented. This is because \textit{TEDM} does not solely rely on contextual information as the \textit{Relabelling Model}, but also on the appearance features extracted by Faster RCNN. Therefore, it is expected to work well even with one object presented as illustrated in Figure \ref{fig: singelObject}, where the predictions obtained from Faster RCNN and \textit{TEDM} are shown on left and right sides, respectively. 

As we can see in the first row in Figure \ref{fig: singelObject}, a \textit{person} is detected by Faster RCNN with a confidence of \textit{0.9985} and also predicted by \textit{TEDM}, but with very slightly lower confidence as \textit{0.9883}. Similarly, in the second row, a \textit{kite} is detected by Faster RCNN as the only object presented. Applying \textit{TEDM}, in this case, increases the confidence from \textit{0.9647} to \textit{0.9920}. The two images are shown as each presents one object that belongs to a different category. In all cases, \textit{TEDM} is believed to work very well when a single object is presented. It can be said that \textit{TEDM} is a compatible method producing comparable results as Faster RCNN. Below, more visualised and statistical results are reported where more than one object in the images used are presented.    

Images with more than one object are used to keep consistent and to report comparable statistical results with the \textit{Relabelling Model}. In Table \ref{Table:New_MethodAUC}, the AUC scores for some objects and all objects average for both \textit{TEDM} and Faster RCNN are presented. It can be noticed that \textit{TEDM} outperforms the baseline detector in all cases shown obtaining a higher average score of \textit{0.8922}. This is even better than the \textit{Relabelling Model}, whose mean of AUC scores is \textit{0.78278}.

Furthermore, Table \ref{Table: maP_NewMEthod} shows the mAP$_0._5$ and mAP
(IoU=[0.5:0.05:0.95]) for \textit{TEDM} and Faster RCNN. As presented, \textit{TEDM} outperforms the baseline. Compared with the \textit{Relabelling Model}, whose mAP and F1 scores are 65.50\% and 58.95\%, as reported with \cite{FaisalAlamri1}, \textit{TEDM} still performs better. 

\begin{table}
\caption{A comparison between Faster RCNN and TEDM in terms of AUC scores of some MS COCO  classes.}\label{Table:New_MethodAUC}
\begin{center}
\begin{tabular}{|c|c|c|c|c|}
\hline
{\cellcolor{yellow!25}Class Name} 
& {\cellcolor{yellow!25}Faster RCNN}  & {\cellcolor{yellow!25}Relabelling Model \cite{FaisalAlamri1}} & {\cellcolor{yellow!25}TEDM}  \\ \hline 
Person  & 0.84345  & 0.84568 & \textbf{0.95174}\\
Car   & 0.81246  & 0.80474 & \textbf{0.93287}\\
Cow   & 0.82301  & 0.84035 & \textbf{0.93246}\\
Snowboard      & 0.74229  & 0.87412 & \textbf{0.88235}\\
Sport ball     & 0.84244  & 0.84203 & \textbf{0.95008}\\
\cellcolor[HTML]{C0C0C0} Mean    & \cellcolor[HTML]{C0C0C0} 0.76472  & \cellcolor[HTML]{C0C0C0} 0.78278 & \cellcolor[HTML]{C0C0C0} \textbf{0.89222} \\ \hline 
\end{tabular}
\end{center}
\end{table}

\begin{table}
\caption{mAP and F1 scores in percentages [\%] for Faster RCNN and Transformer-Encoder Detector Module (TEDM). }\label{Table: maP_NewMEthod}
\begin{center}
\begin{tabular}{|c|c|c|c|}
\hline
{\cellcolor{gray!25} Model} & {\cellcolor{gray!25} mAP$_0._5$} & {\cellcolor{gray!25} mAP} & {\cellcolor{gray!25} F1}   \\\hline 
{\cellcolor{yellow!25} Faster RCNN} & 62.82 & 33.48 & 57.34 \\\hline 
{\cellcolor{yellow!25} Relabelling Model \cite{FaisalAlamri1}} & 65.50 & 34.07 & 58.95 \\\hline 
{\cellcolor{yellow!25}Transformer-Encoder Detector Module} & \textbf{69.07}  & \textbf{38.19}&  \textbf{63.27} \\\hline 
\end{tabular}
\end{center}
\end{table}

\begin{figure*}
\begin{center} 

\begin{tabular}{|c|c|c|}  \hline 
{\color{blue}{\cellcolor{green!25}Original Image}} & {\color{blue}{\cellcolor{green!25}Faster RCNN}}  & {\color{blue}{\cellcolor{green!25}TEDM}} \\ \hline
\includegraphics[width=0.20\linewidth]{}
& 
\includegraphics[width=0.20\linewidth]{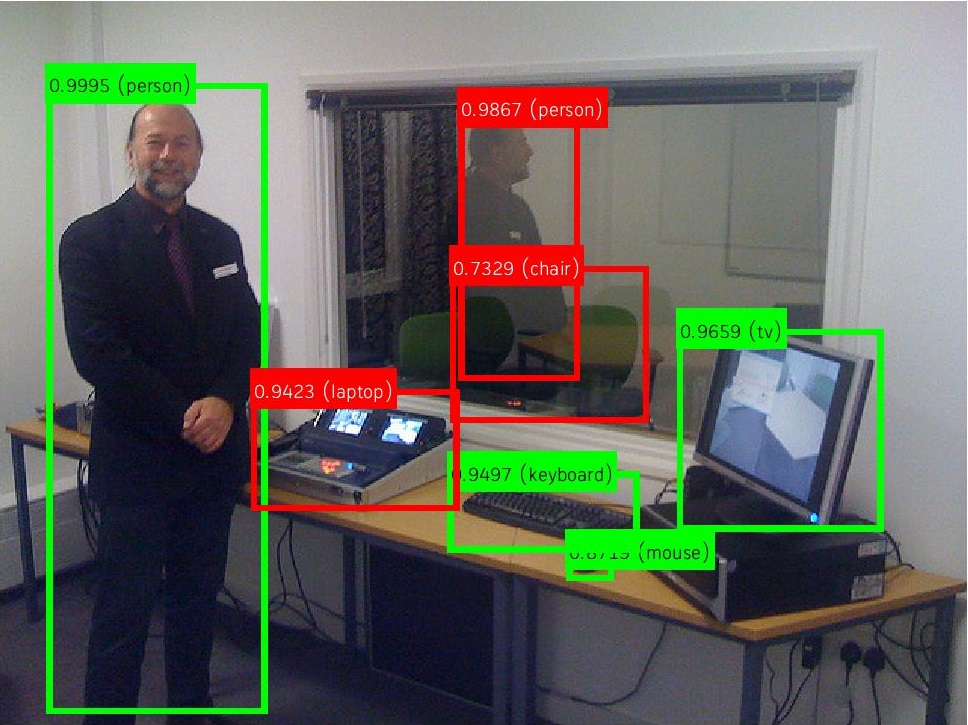}
& 
\includegraphics[width=0.20\linewidth]{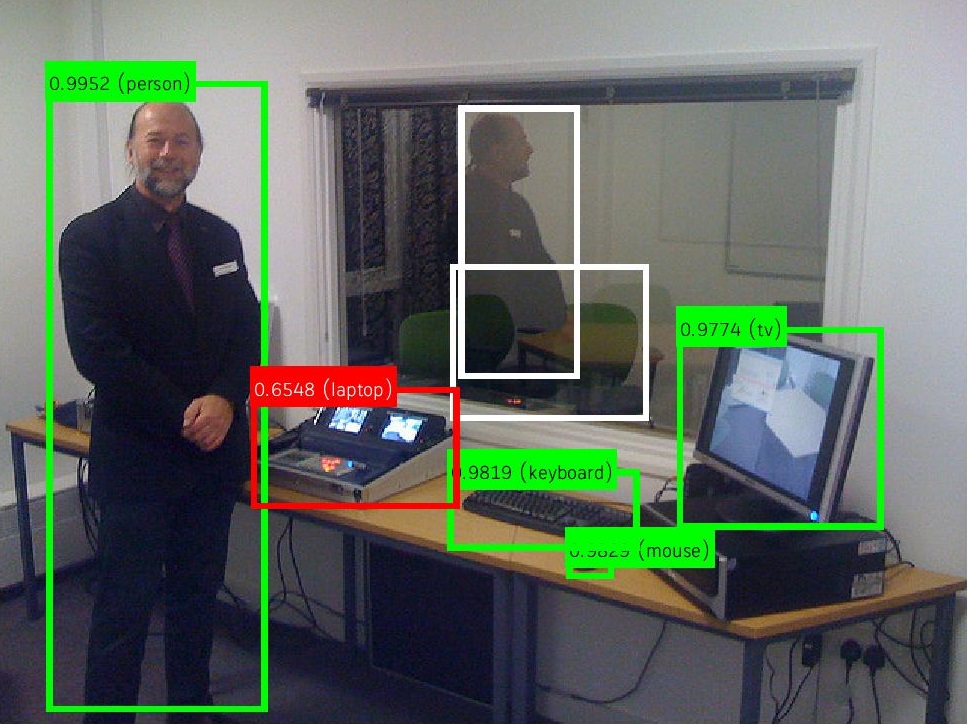}
\\ \hline

\includegraphics[width=0.20\linewidth]{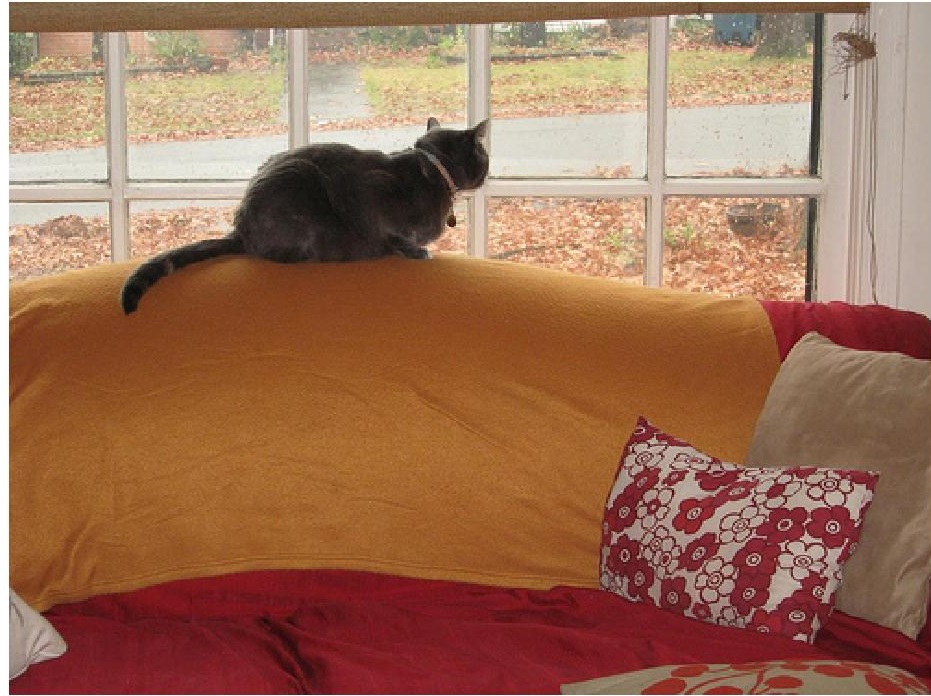}
& 
\includegraphics[width=0.20\linewidth]{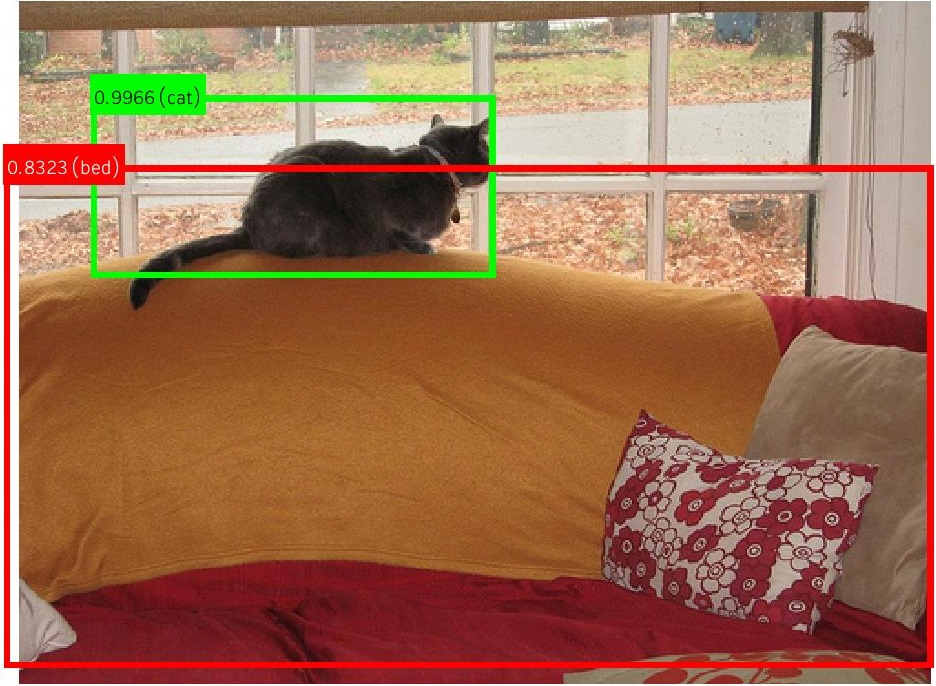}
& 
\includegraphics[width=0.20\linewidth]{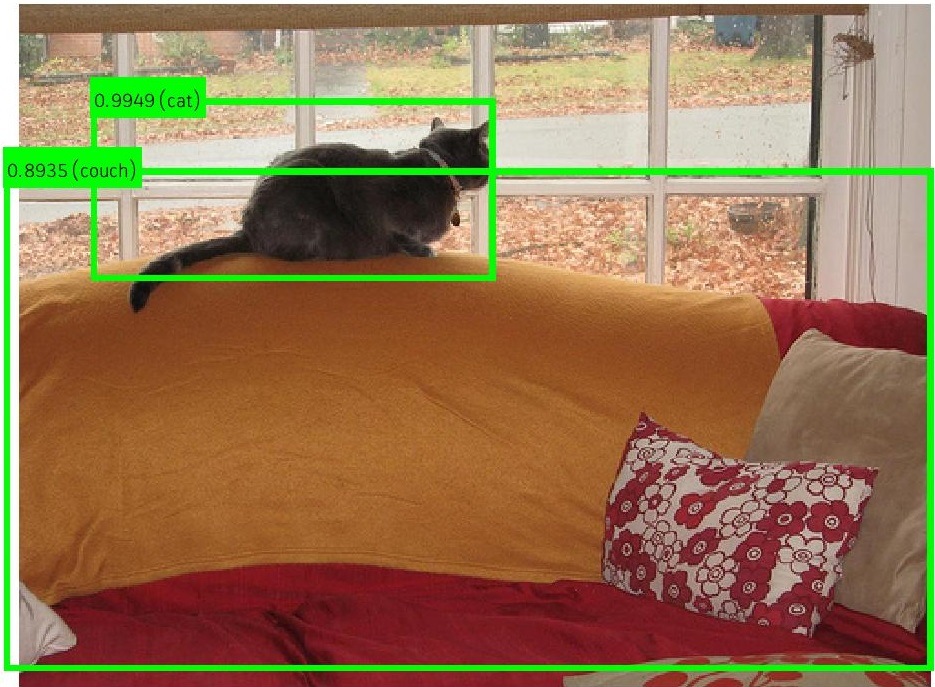}
\\ \hline

\end{tabular}
\end{center}
\caption{Results: Faster RCNN vs. TEDM outputs: Green, red and white boxes represent correct detection, incorrect detection, and objects removed and relabelled as background, respectively}
\label{fig: NoAttackTEDM}
\end{figure*}

In addition, as shown in Figure \ref{fig: NoAttackTEDM}, two randomly chosen images taken from MS COCO 2017 validation dataset are used to examine the performance of \textit{TEDM} compared with Faster RCNN. 
In the first row, seven objects are detected by Faster RCNN. Four objects, which are a \textit{person}, a \textit{TV}, a \textit{keyboard} and a \textit{mouse} are detected correctly. However, the other three, which are a \textit{person}, a \textit{laptop} and \textit{a chair} are incorrectly detected.  \textit{TEDM} is applied, which removes two of the incorrectly detected objects, but still predicts the laptop incorrectly, with very low confidence. This result is promising as it shows how \textit{TEDM} removes incorrect objects and reduces the confidence of the \textit{laptop}. More importantly, in the following image, \textit{TEDM} outperforms Faster RCNN predictions. It corrects the \textit{bed}, which is incorrectly detected, to a \textit{couch} as correct detection. This clearly shows how the proposed model does not only contribute to removing false detections but also how it rescores and relabels them.

\textbf{Summary}. Statistical and visual results suggest that  \textit{TEDM} achieves a better performance than the baseline detector benefiting from the self-attention mechanism. It also does a great job in detecting either single-object or multiple objects, which are better in comparison with Faster RCNN and the \textit{Relabelling Model} \cite{FaisalAlamri2}.

\subsection{Experiment Two: Adversarial Images} \label{AttackSection}
Adversarial images are used to examine the impact of \textit{TEDM}, i.e. the use of contextual information, against adversarial attacks. Such images may have different  visual features due to the addition of Adversarial Perturbations leading to an effect on contextual information especially when some objects are misdetected. 

Adversarial Perturbations are noises carefully computed such that, when added to images, they are not visible to the naked human eye but they do fool the Deep Neural Networks (DNNs) into mispredictions.

\begin{figure}
\begin{center} 

\begin{tabular} {|c|c|}  \hline 
{\textbf{\color{blue}{\cellcolor{red!25}FFF Perturbations}}}  & {\textbf{\color{blue}{\cellcolor{green!25}UAP Perturbations}}} \\ \hline
\includegraphics[width=0.25\linewidth]{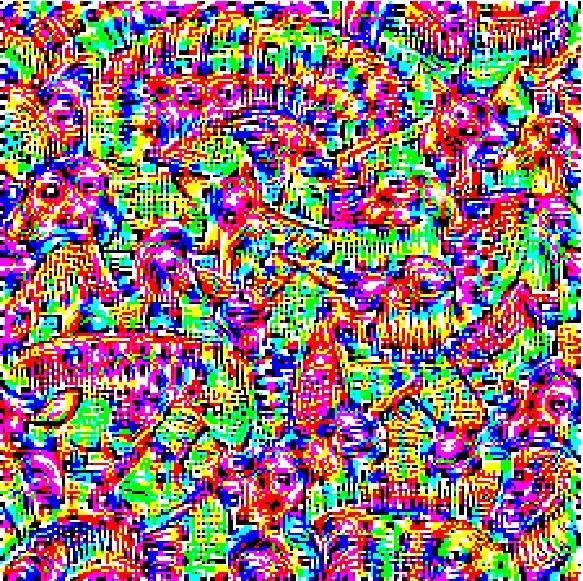} &
\includegraphics[width=0.25\linewidth]{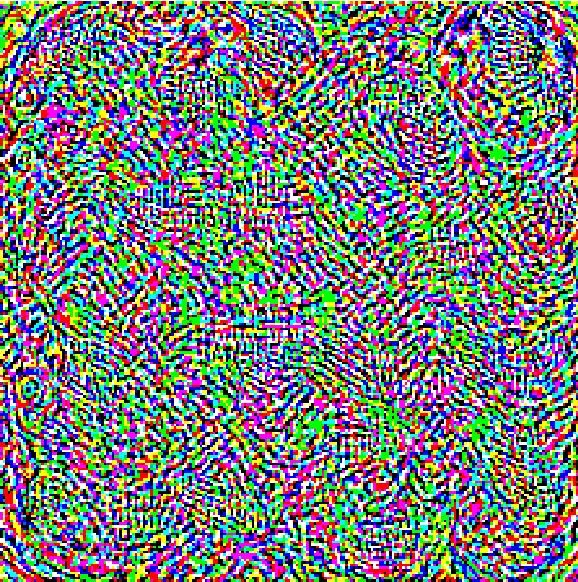}  \\ \hline
\end{tabular}
\end{center}
\caption{FFF vs. UAP perturbations added to images}
\label{fig:perturbationsFFF_UAP}
\end{figure}

Due to the impact that both methods (i.e. UAP and FFF) cause to the CNN models, they are used in this paper to examine their effect on \textit{TEDM} and Faster RCNN detector. UAP perturbation used in this experiment is trained on MS COCO 2017, as it is data-dependent. However, FFF is used as proposed (i.e. already trained on ImageNet), but will be tested on MS COCO 2017. Figure \ref{fig:perturbationsFFF_UAP} shows the perturbations, which are added on images. It can be seen they are different, even though they both are trained on the same model (i.e. VGG-16), but on different training datasets.

The two types of perturbations are applied and added to images in two different approaches: (1) added to the entire image as experimented in Section \ref{EXEntireImage}, (2) added to each detected region as experimented in Section \ref{EXROIImage}. 

\subsubsection{Perturbation on the Entire Image} \label{EXEntireImage}
In this section, perturbations are added to the entire image, where statistical and visual results are illustrated, below, to show the performance of \textit{TEDM} in comparison with Faster RCNN.

In terms of the statistical results, as shown in Table \ref{Table:Entire_Image_AUC_TEDM}. The average AUC scores for \textit{TEDM} is noticeably higher than Faster RCNN in both types of attacks. Considerably, \textit{TEDM} obtains higher scores for some objects such as the \textit{train}, \textit{stop sign} and \textit{Pizza}, we can clearly see a huge difference, as \textit{TEDM} benefits from the use of visual and contextual features in contrast to Faster RCNN, which depends only on the visual features. In comparison with the \textit{Relabelling Model}, which encodes contextual information in an explicitly manner, the average AUC scores are \textit{0.4203}, and \textit{0.4333} for FFF and UAP attacks respectively.   

In addition, Table \ref{Table:EntireImageTEDM} illustrates the mAP and F1 scores for \textit{TEDM} and Faster RCNN, where FFF and UAP perturbations are applied. Clearly, \textit{TEDM} scores a higher performance than Faster RCNN in both cases. However, in the case of UAP attack, we can see that both models are less impacted compared with FFF. In comparison with the \textit{Relabelling Model} which obtains mAP and F1 of 49.39\%, 41.91\% and 50.72\%, 43.12\% for FFF and UAP respectively.

\begin{table}
\caption{A comparison between some MSCOCO object classes AUC scores between Faster RCNN and TEDM on perturbed images}\label{Table:Entire_Image_AUC_TEDM}
\begin{center}
\begin{tabular}{|c|c|c|c|c|}
\hline
\multicolumn{1}{|c|}{{\cellcolor{yellow!25}}}  &
\multicolumn{2}{|c|}{\textbf{\color{blue}{\cellcolor{red!25} FFF }}}  &
\multicolumn{2}{|c|}{\textbf{\color{blue}{\cellcolor{green!25} UAP}}}  \\ 

 {\cellcolor{yellow!25}Class Label} & {\cellcolor{red!25}Detector} & {\cellcolor{red!25}TEDM} & {\cellcolor{green!25}Detector} & {\cellcolor{green!25}TEDM} \\ \hline 
Person & {\cellcolor{red!25}0.62389} & {\cellcolor{red!25}\textbf{0.69576}} & {\cellcolor{green!25}0.62881} & {\cellcolor{green!25}\textbf{0.73342}}  \\ 


Train & {\cellcolor{red!25}0.42961} & {\cellcolor{red!25}\textbf{0.62051}} & {\cellcolor{green!25}0.41062} & {\cellcolor{green!25}\textbf{0.56773}}  \\ 

Stop Sign & {\cellcolor{red!25}0.76242} & {\cellcolor{red!25}\textbf{1}} & {\cellcolor{green!25}0.69155} & {\cellcolor{green!25}\textbf{0.78571}}  \\ 



Surfboard & {\cellcolor{red!25}0.39266} & {\cellcolor{red!25}\textbf{0.52464}} & {\cellcolor{green!25}0.44750} & {\cellcolor{green!25}\textbf{0.59365}}  \\ 


Pizza & {\cellcolor{red!25}0.67393} & {\cellcolor{red!25}\textbf{0.87767}} & {\cellcolor{green!25}0.63185} & {\cellcolor{green!25}\textbf{0.81783}}  \\ 


\cellcolor[HTML]{C0C0C0} Mean  & {\cellcolor{red!25}0.34714} & {\cellcolor{red!25} \textbf{0.47683}} & {\cellcolor{green!25} 0.36050} & {\cellcolor{green!25} \textbf{0.48713}}  \\ \hline 
\end{tabular}
\end{center}
\end{table}

\begin{table}
\caption{AP and F1 scores in percentages [\%] for Faster RCNN and TEDM on perturbed images}\label{Table:EntireImageTEDM}
\begin{center}
\begin{tabular}{|c|c|c|c|}
\hline
 {\cellcolor{gray!25}Attacks } & {\cellcolor{gray!25}Models} &  {\cellcolor{gray!25}  \space \space \space mAP$_{0._5}$}  \space \space \space &  {\cellcolor{gray!25}  \space \space \space F1 Score}  \\ \hline

\multirow{ 2}{*}{{\color{red}FFF}}  & {\cellcolor{red!25} Faster RCNN } & {\cellcolor{red!25}45.34\%}  & {\cellcolor{red!25}39.13\%} \\ 

& {\cellcolor{red!25} TEDM }  & {\cellcolor{red!25}\textbf{50.59\%}} & {\cellcolor{red!25}\textbf{43.16\%}} \\ \hline 


\multirow{ 2}{*}{{\color{green}UAP}}  & {\cellcolor{green!25} Faster RCNN } & {\cellcolor{green!25}46.40\%}  & {\cellcolor{green!25}40.05\%}  \\

& {\cellcolor{green!25} TEDM }   & {\cellcolor{green!25}\textbf{51.67\%}} & {\cellcolor{green!25}\textbf{44.00\%}}  \\ \hline

\end{tabular}
\end{center}
\end{table}

\begin{figure*}

\begin{center} 

\begin{tabular}{|c|c|c|c|}  \hline 
{\cellcolor{yellow!25}Perturbation} & {\cellcolor{yellow!25}Original Image} & {\cellcolor{yellow!25}Faster RCNN}  & {\cellcolor{yellow!25}TEDM} \\ \hline
{\textbf{\cellcolor{red!25}FFF }}&
\rowincludegraphics[width=0.20\linewidth]{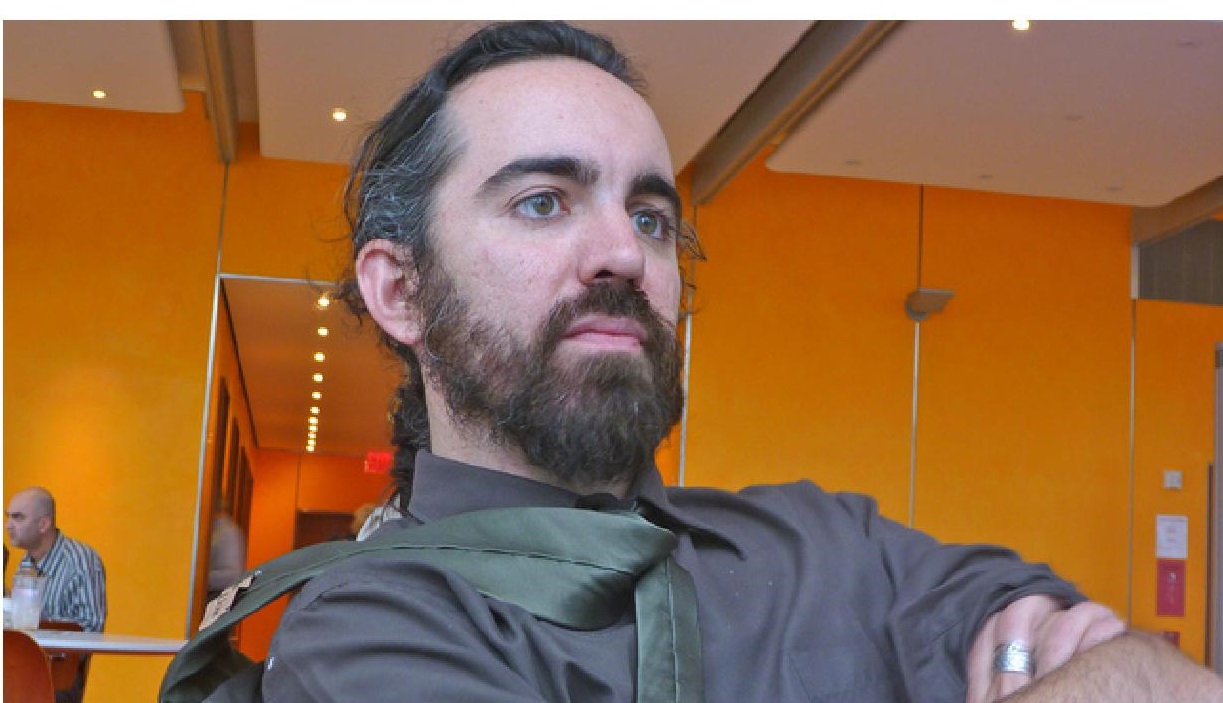}
& 
\rowincludegraphics[width=0.20\linewidth]{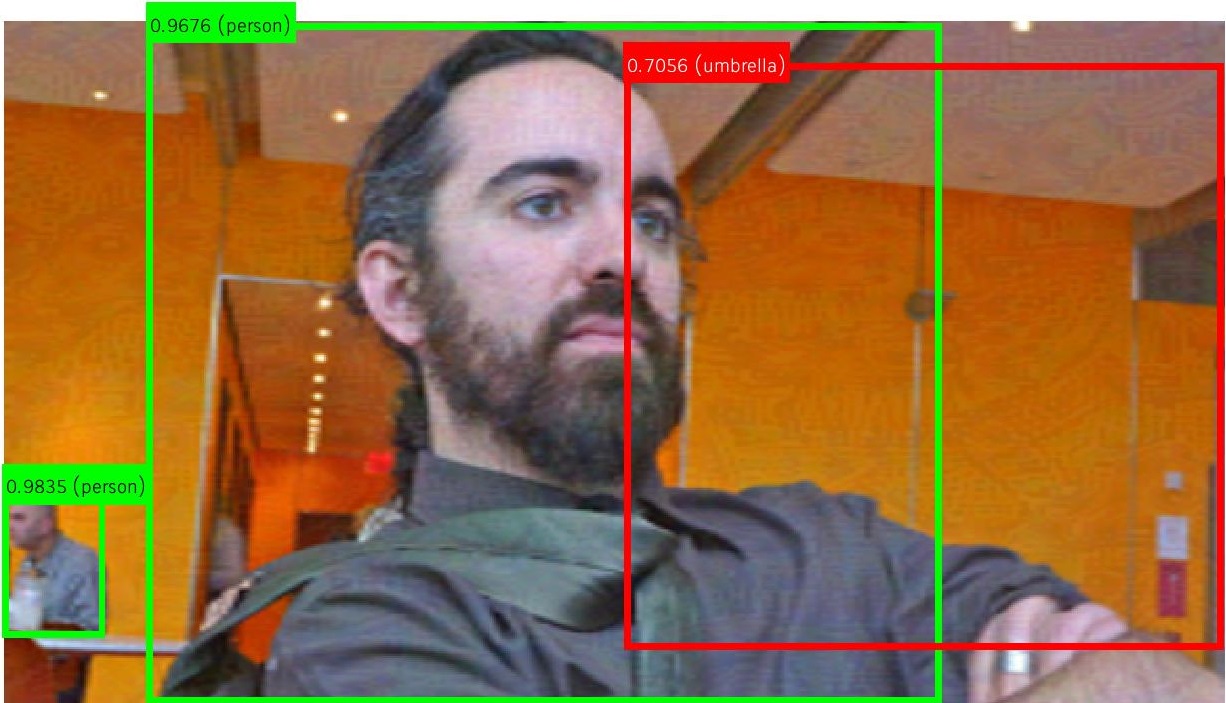}
& 
\rowincludegraphics[width=0.20\linewidth]{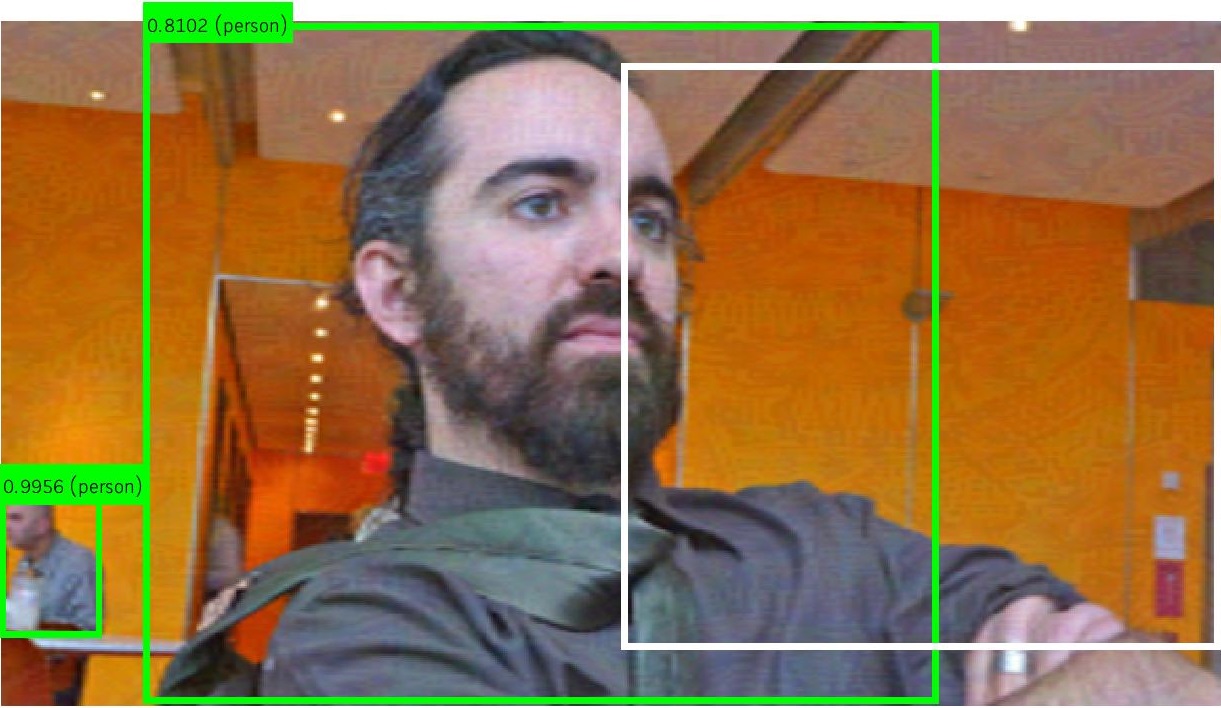}
\\ \hline

{\textbf{\cellcolor{green!25}UAP }} & 
\rowincludegraphics[width=0.20\linewidth]{}
& 
\rowincludegraphics[width=0.20\linewidth]{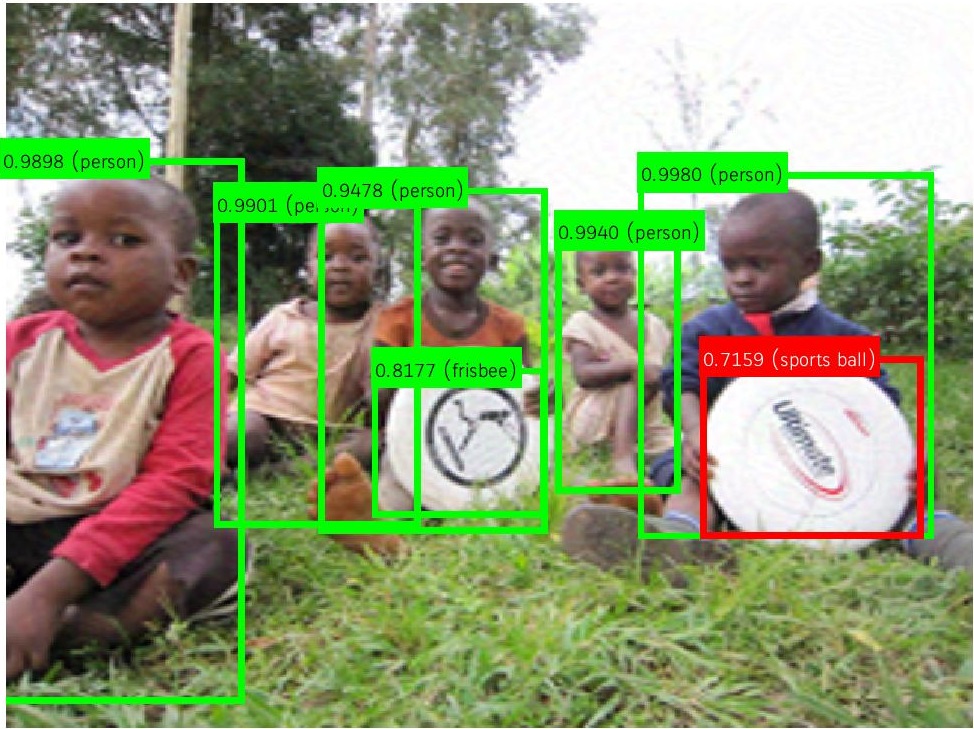}
& 
\rowincludegraphics[width=0.20\linewidth]{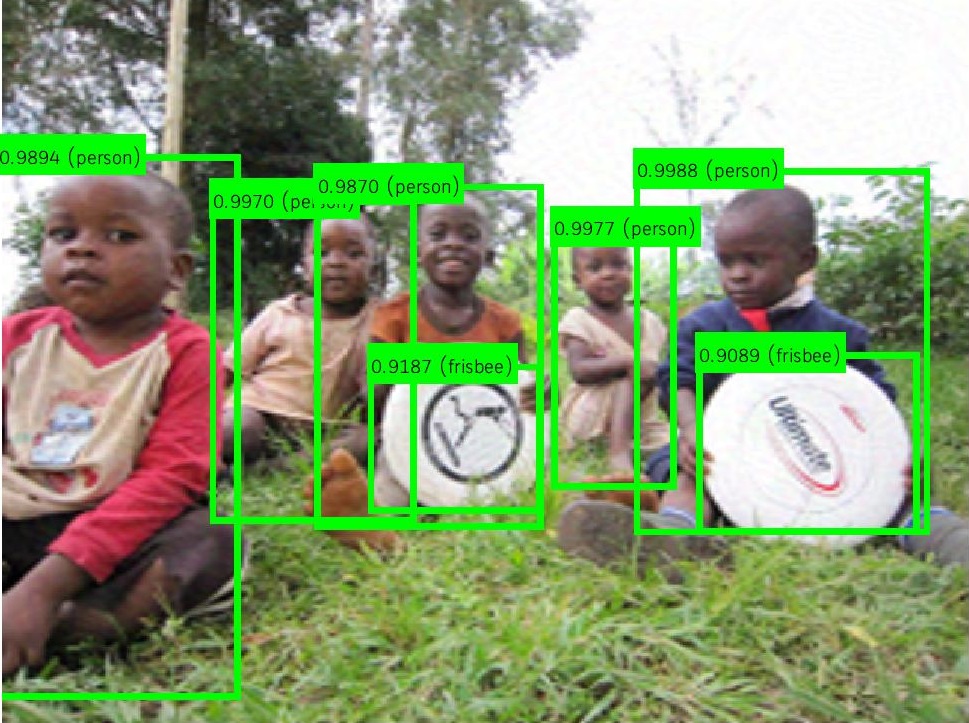}
\\ \hline

\end{tabular}
\end{center}
\caption{Results: Faster RCNN and TEDM outputs for FFF and UAP perturbed images: Green, red and white boxes represent correct detection, incorrect detection, and objects removed and relabelled as background, respectively}
\label{fig:EntireImageTEDM}
\end{figure*}

Figure \ref{fig:EntireImageTEDM} presents two images illustrating the impact of FFF and UAP perturbations. Firstly, in the top row results, FFF perturbation is added to the entire image. Faster RCNN detects two objects correctly, which are two \textit{persons}, whereas incorrectly detects an \textit{umbrella}, which is actually a \textit{wall}. The same perturbed image is passed into the \textit{TEDM} predicting the presence of only the correct objects, which are the two \textit{persons}, and relabels the \textit{umbrella} as a background. However, in compare with Faster RCNN, \textit{TEDM} scores the correctly detected objects with lower confidences, but still higher than \textit{0.81}.

In terms of UAP perturbations, as illustrated in the second row. We can see that both models perform very well. However, Faster RCNN fails to detect one of the \textit{frisbees}, which labels as a \textit{sports ball}. \textit{TEDM} correctly labels it as \textit{frisbee} with a confidence of \textit{0.9089}. Both models score all other objects with considerably high confidences, even though perturbations are added. 

\textbf{Summary}. We can clearly see that \textit{TEDM} performs better scoring higher average AUC scores in the majority of cases, and higher mAP and F1 scores compared with Faster RCNN and the \textit{Relabelling Model}. From the reported results, we can say that \textit{TEDM} is an excellent tool to overcome some of the errors that Faster RCNN attempts. Below, both models are examined when perturbations are added on regions rather than on the entire image.

\subsubsection{Perturbation On Regions} \label{EXROIImage}
In this section, perturbation is added on detected regions, because perturbations are added to images, which are passed into single-object models, as presented and developed in \cite{MainPaper_Ch5, FastFeatureFool}. Therefore, perturbations will be added to each region of interest (RoI) that the detector detects during the detection process.

Looking at Table \ref{Table:MAPF1TEDM} we can see that \textit{TEDM} outperforms Faster RCNN in both types of perturbation. FFF seems to impact both models lower than what UAP does, but in general, both attacks significantly drop the performance of both models. We believe \textit{TEDM} can be offered, to some extent, yet as a solution to address perturbation issue in the field of object detection. It also outperforms the \textit{Relabelling Model}, whose mAP and F1 scores for both FFF and UAP perturbations are respectively 29.05\%, 23.14\% and 26.36\%, 20.53\%. 

AUC scores are computed as presented in Table \ref{Table:ROIImage_AUC_TEDM}. On average, \textit{TEDM} performs better than Faster RCNN in the two cases of perturbation. As said, both models perform better when FFF attack is applied, unlike when UAP perturbation added. However, in some classes such as \textit{microwave} Faster RCNN has higher AUC scores (as per class) comparing with the proposed method. In comparison with the \textit{Relabelling Model}, which has an average AUC score for FFF and UAP as 0.19470 and 0.16994, where \textit{TEDM} again achieves better performance.

\begin{table}
\caption{AP and F1 scores in percentages [\%] for Faster RCNN and TEDM on perturbed images}\label{Table:MAPF1TEDM}
\begin{center}
\begin{tabular}{|c|c|c|c|}
\hline
 {\cellcolor{gray!25}Attacks } & {\cellcolor{gray!25}Models} &  {\cellcolor{gray!25}  \space \space \space mAP$_{0.5}$}  \space \space \space &  {\cellcolor{gray!25}  \space \space \space F1 Score}  \\ \hline

\multirow{ 2}{*}{{\color{red}FFF}}  & {\cellcolor{red!25} Faster RCNN } & {\cellcolor{red!25}26.25\%}  & {\cellcolor{red!25}20.78\%} \\ 

& {\cellcolor{red!25} TEDM }  & {\cellcolor{red!25}\textbf{35.28\%}} & {\cellcolor{red!25}\textbf{26.25\%}} \\ \hline 


\multirow{ 2}{*}{{\color{green}UAP}}  & {\cellcolor{green!25} Faster RCNN } & {\cellcolor{green!25}23.18\%}  & {\cellcolor{green!25}18.17\%}  \\

& {\cellcolor{green!25} TEDM }   & {\cellcolor{green!25}\textbf{31.39\%}} & {\cellcolor{green!25}\textbf{24.14\%}}  \\ \hline

\end{tabular}
\end{center}
\end{table}

\begin{table}
\caption{A comparison between some MSCOCO object classes AUC scores between detector and TEDM on perturbed Regions}\label{Table:ROIImage_AUC_TEDM}
\begin{center}
\begin{tabular}{|c|c|c|c|c|}
\hline
\multicolumn{1}{|c|}{{\cellcolor{yellow!25}}}  &
\multicolumn{2}{|c|}{\textbf{\color{blue}{\cellcolor{red!25} FFF }}}  &
\multicolumn{2}{|c|}{\textbf{\color{blue}{\cellcolor{green!25} UAP}}}  \\ 

 {\cellcolor{yellow!25}Class Label} & {\cellcolor{red!25}Detector} & {\cellcolor{red!25}TEDM} & {\cellcolor{green!25}Detector} & {\cellcolor{green!25}TEDM} \\ \hline 
Person & {\cellcolor{red!25}0.18573} & {\cellcolor{red!25}\textbf{0.31982}} 
& {\cellcolor{green!25}0.14653} & {\cellcolor{green!25}\textbf{0.28385}}  \\ 


Train & {\cellcolor{red!25}0.05501} & {\cellcolor{red!25}\textbf{0.08737}} 
& {\cellcolor{green!25}\textbf{0.03883}} & {\cellcolor{green!25}0.02912}  \\ 

Cat & {\cellcolor{red!25}0.06807} & {\cellcolor{red!25}\textbf{0.10563}} 
& {\cellcolor{green!25}0.03990} & {\cellcolor{green!25}\textbf{0.84507}}  \\ 




Wine Glass & {\cellcolor{red!25}\textbf{0.27083}} & {\cellcolor{red!25}0.19791 }
& {\cellcolor{green!25}0.24791} & {\cellcolor{green!25}\textbf{0.25625}}  \\ 


Microwave & {\cellcolor{red!25}\textbf{0.17241}} & {\cellcolor{red!25}0 }
& {\cellcolor{green!25}0.06896} & {\cellcolor{green!25}\textbf{0.09143}}  \\ 


\cellcolor[HTML]{C0C0C0} \textbf{Mean}  & \cellcolor[HTML]{C0C0C0} 0.14310 & \cellcolor[HTML]{C0C0C0}\textbf{0.23058} & \cellcolor[HTML]{C0C0C0}0.12043 & \cellcolor[HTML]{C0C0C0}\textbf{0.19991}  \\ \hline 
\end{tabular}
\end{center}
\end{table}

\begin{figure*}
\begin{center} 

\begin{tabular}{|c|c|c|c|}  \hline 
{\cellcolor{yellow!25}Perturbation} & {\cellcolor{yellow!25} Regions Before Attack} & {\cellcolor{yellow!25}Faster RCNN}  & {\cellcolor{yellow!25}TEDM} \\ \hline
{\textbf{\cellcolor{red!25}FFF }} &
\rowincludegraphics[width=0.20\linewidth]{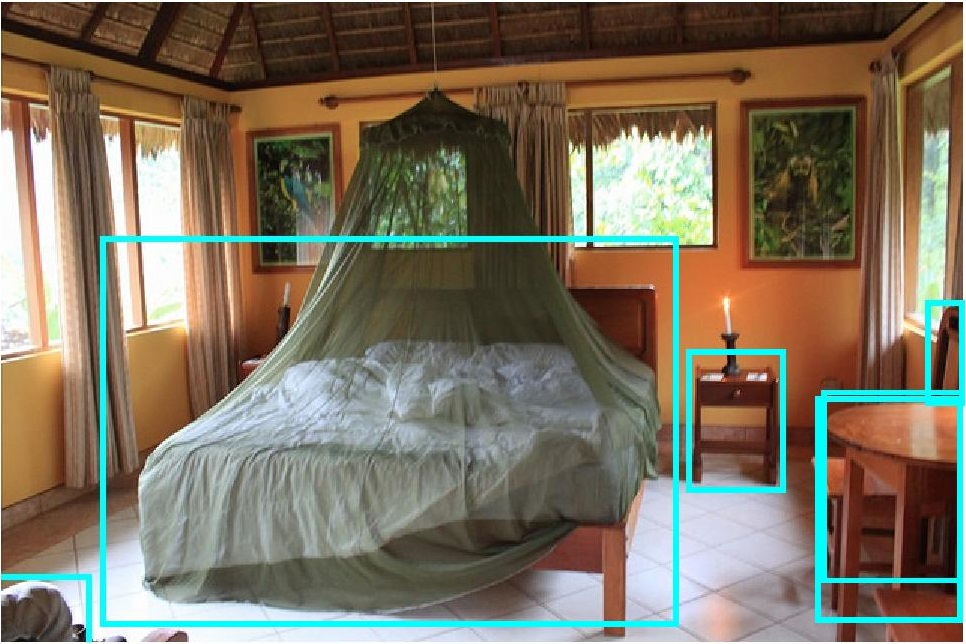}
& 
\rowincludegraphics[width=0.20\linewidth]{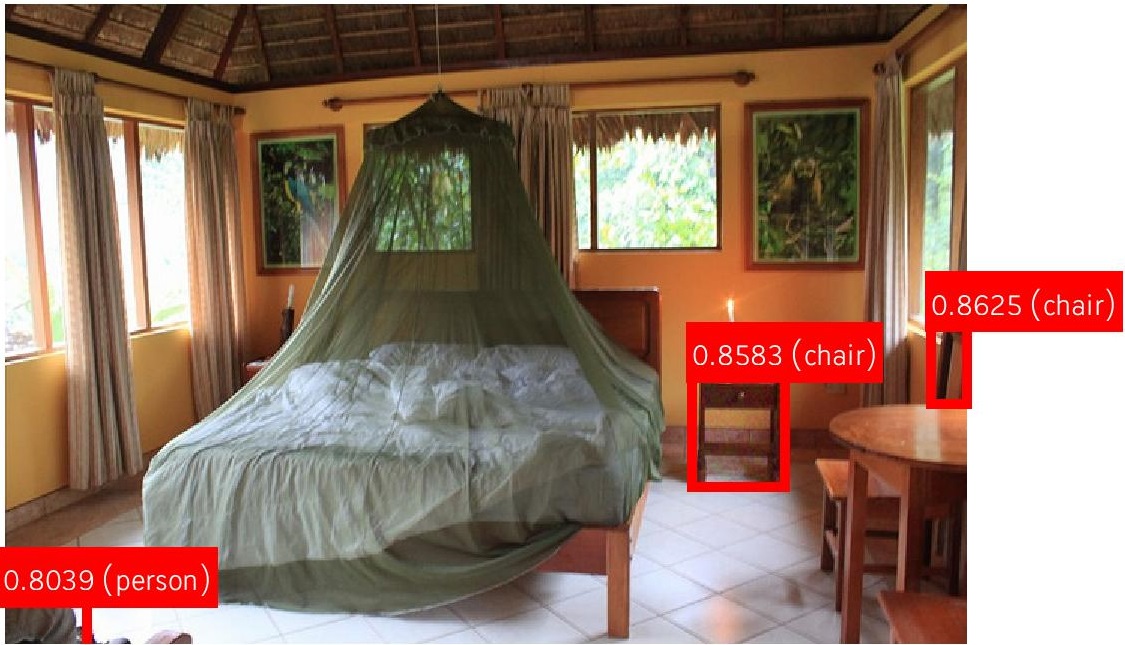}
& 
\rowincludegraphics[width=0.20\linewidth]{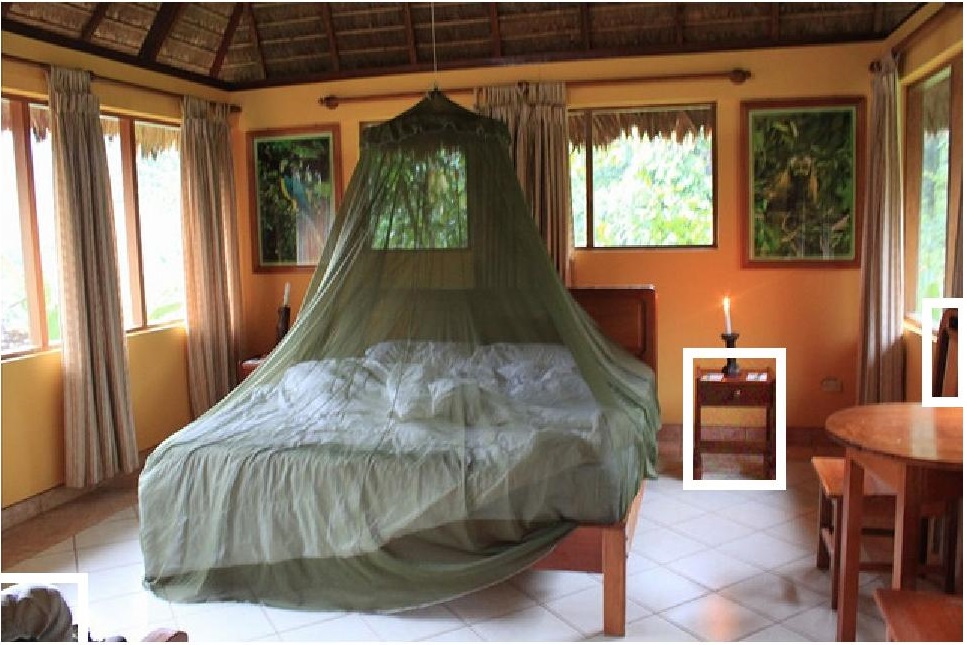}
\\ \hline

{\textbf{\cellcolor{green!25}UAP }} &
\rowincludegraphics[width=0.20\linewidth]{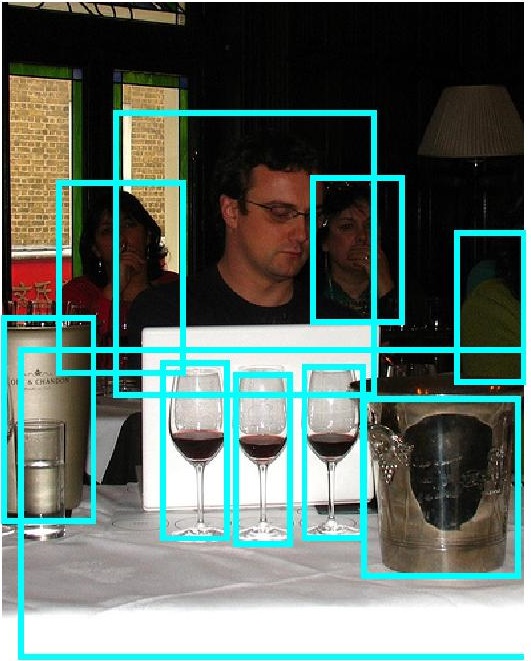}
& 
\rowincludegraphics[width=0.20\linewidth]{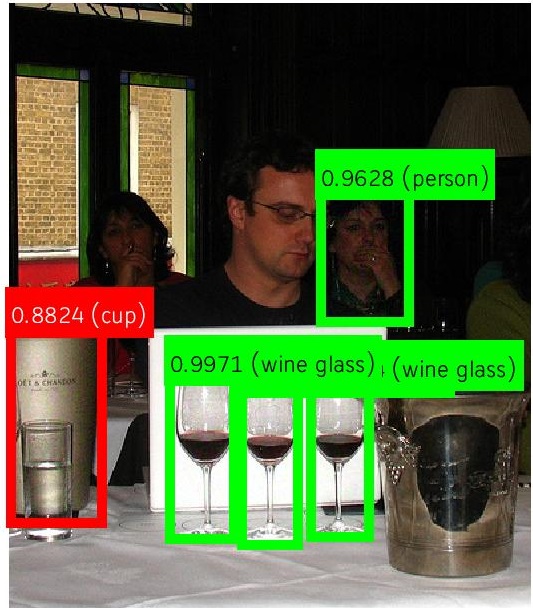}
& 
\rowincludegraphics[width=0.20\linewidth]{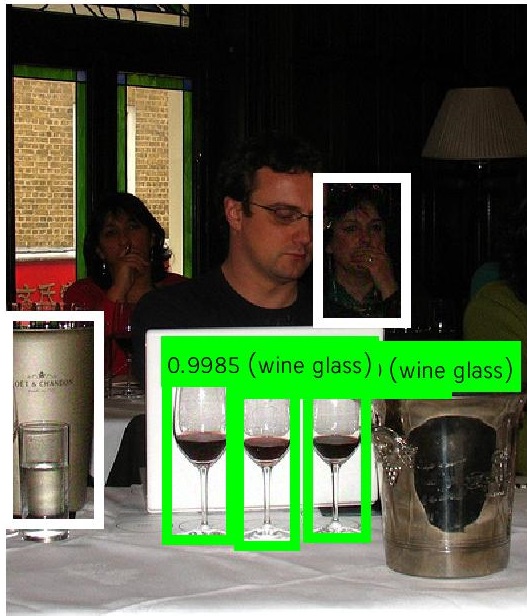}
\\ \hline

\end{tabular}
\end{center}
\caption{Results: Faster RCNN and TEDM outputs for FFF and UAP perturbed regions: Light blue boxes represent regions detected before perturbation added. Green, red and white boxes represent correct detection, incorrect detection, and objects removed and relabelled as background, respectively. }
\label{fig:TEDMRoI_Image_Results_FFF_UAP}
\end{figure*}

Table \ref{fig:TEDMRoI_Image_Results_FFF_UAP} presents some results obtained from the application of both attacks on Faster RCNN and \textit{TEDM}. An image showing the detected regions before any perturbation added, followed by the predictions of both models are illustrated. As the aim, here, is not just to illustrate the differences between models performances before and after the attack, but rather to show the differences among both models after attacked per region is added. Therefore, results after the attack are reported to provide ease when comparing. 

In the first image, where FFF perturbation is added to regions \textit{TEDM} predicts no objects. All objects detected by Faster RCNN after attacked are false detection. \textit{TEDM} helps to prevent false detections that Faster RCNN outputs. Noticeably, as shown in the detected regions before the attack, the \textit{bed} is detected, but after the attack, it could not be predicted. This is a good example of how negatively the perturbation impacts the model. It is found that when the regions are large in size, they are likely to be impacted more by perturbation resulting in not being detected. 

In terms of UAP perturbation, ten regions are detected by Faster RCNN before the attack, and only half of them are detected after. Faster RCNN detects four objects correctly but fails in detects the \textit{cup}. The \textit{TEDM} fails to detect the \textit{person} that Faster RCNN already detects, which can be due to the perturbation and lighting conditions. We can see that the \textit{person} in the large region is not detected, which can be in line with the findings stated earlier that larger regions are more likely to be impacted.

\textbf{Summary}. \textit{TEDM} is performing better than Faster RCNN to tackle adversarial perturbations. This is because of the features encoding during the encoder process, as it learns the spatial and visual features.   

\section{Conclusion}
This paper presented a new context module, called \textit{Transformer-Encoder Detector Module (TEDM)}, which when combined with an object detection architecture to improve both performance and robustness to adversarial attacks. The proposed model is based on the encoder part from the Transform model \cite{Attention:journals/corr/VaswaniSPUJGKP17} to encode the contextual features implicitly for scenes, and visual features using Faster RCNN. \textit{TEDM} was examined on natural images and perturbed images, where it outperforms Faster RCNN and a contextual model that explicitly encodes contextual features.
 
As experimented, the impact of adversarial attacks was reported to be higher when applied on regions, which we believe is due to the size of the regions: the larger the region is, the more it is impacted. Surprisingly, UAP perturbation affects the performance of the examined models when added to the entire image less than FFF does. However, when added to regions, it drops performances considerably, which can be due to the data-dependency.

Future work will involve developing an end-to-end model, to refine not only predictions but also boundary boxes from both contextual and visual features. 

\bibliographystyle{plain}
\bibliography{malliopas}

\end{document}